\documentclass{article}
\usepackage{multirow}
\usepackage{hyperref}
\usepackage{spconf,amsmath,graphicx, hyperref,amsfonts, adjustbox}
\usepackage{booktabs}
\usepackage[table,xcdraw]{xcolor}

\title{Low-rank Adaptation of Large Language Model Rescoring for \\Parameter-Efficient Speech Recognition}
\name{Yu Yu\thanks{$^*$Work done as an applied scientist intern at Amazon Alexa.}, Huck Yang, Jari Kolehmainen, Prashanth G. Shivakumar, Yile Gu, Sungho Ryu, Roger Ren, Qi Luo, Aditya Gourav, I-Fan Chen, Yi-Chieh Liu, Tuan Dinh, Ankur Gandhe, Denis Filimonov, Shalini Ghosh, Andreas Stolcke, Ariya Rastow, Ivan Bulyko}
\name{%
\begin{tabular}{@{}c@{}}Yu Yu$^*$\thanks{$^*$Work done as an applied scientist intern at Amazon Alexa.}, Chao-Han Huck Yang, Jari Kolehmainen, Prashanth G. Shivakumar, Yile Gu, Sungho Ryu \\ Roger Ren, Qi Luo, Aditya Gourav, I-Fan Chen, Yi-Chieh Liu, Tuan Dinh, Ankur Gandhe \\ Denis Filimonov, Shalini Ghosh, Andreas Stolcke, Ariya Rastow, Ivan Bulyko
\end{tabular}}
\address{Amazon, USA \\ $^*$Stevens Institute of Technology, USA }
\begin{document}
\maketitle
\begin{abstract}
We propose a neural language modeling system based on low-rank adaptation (LoRA) for speech recognition output rescoring. Although pretrained  language models (LMs) like BERT have shown superior performance in second-pass rescoring, the high computational cost of scaling up the pretraining stage and adapting the pretrained models to specific domains limit their practical use in rescoring. Here we present a method based on low-rank decomposition to train a rescoring BERT model and adapt it to new domains using only a fraction (0.08\%) of the pretrained parameters. These inserted matrices are optimized through a discriminative training objective along with a correlation-based regularization loss. The proposed low-rank adaptation RescoreBERT (LoRB) architecture is evaluated on LibriSpeech and internal datasets with decreased training times by factors between 5.4 and 3.6.

\end{abstract}
\section{Introduction}

Second-pass rescoring is a widely explored technique to improve the performance of automatic speech recognition (ASR) systems~\cite{gaur2022multilingual,hu2021transformer,gandhe2020audio,sainath2019two, hung2023low}. Language models in different architectures, such as long short-term memory (LSTM)~\cite{hochreiter1997long} and
transformer~\cite{vaswani2017attention}, have proven effective as N-best rescorers~\cite{yang2021multi} to boost the performance of first-pass decoding. Notably, transformers stand out among other language model architectures due to their exceptional ability to model long-range
dependencies and context within the input. Additionally, large language models (LLMs) such as GPT-2~\cite{zheng2021adapting} and BERT~\cite{shin2019effective}, which are based on transformers, have the advantage of incorporating both linguistic and world knowledge. As a result, LLMs have been used in extensive applications across many natural language processing tasks.  

LLMs are conventionally pretrained on massive unlabelled data sets and fine-tuned on some smaller labelled datasets for adaptation to downstream tasks. However, as the size of the pretrained models increases, the cost associated with fine-tuning and deploying these models for real-world applications also escalates. To address this practical challenge, a range of parameter-efficient methods (e.g., adapters, model reprogramming, and prompts) have been proposed~\cite{li2021prefix,houlsby2019parameter,zaken2021bitfit, yang2023english, yen2021neural, ho2023differentially, chang2022speechprompt, chang2023speechprompt} to alleviate the computation and memory demands of fine-tuning LLMs. Low-rank adaptation (\textbf{LoRA})~\cite{hu2021lora} freezes all pretrained parameters in the LLM and inserts a trainable pair of matrices (acting as a low-rank decomposition of a full matrix) additively into each layer of the Transformer architecture. Compared to other parameter-efficient training methods, such as adapters~\cite{houlsby2019parameter}, LoRA has two distinct advantages: 1) it employs a simple architecture and has the potential to reduce the number of trainable parameters compared to alternatives; 2) LoRA does not introduce any additional inference latency, making it an excellent choice for deployment in production environments.

In this work, we explore low-rank adaptation for language model rescoring 
to achieve a favorable trade-off between computational efficiency and speech recognition performance. Specifically, we follow the discriminative training objective proposed in \cite{xu2022rescorebert} to directly optimize the minimum word error rate, as described in Section~\ref{sec:mwer}. During training, we freeze all layers in BERT and only update low-rank matrices inserted at each transformer layer, as discussed in Section~\ref{sec:lora}. As a result, the memory required to store the trainable parameters and the backward-pass computation are both reduced. Meanwhile, it is worth noting that we have observed that LoRA can lead to a degraded representation, similar to full fine-tuning~\cite{jiang2019smart}, which can consequently affect performance on unseen test domains. To mitigate this negative effect, we further apply a correlation-based regularization in addition to the minimum word error loss, as shown in Section~\ref{sec:multiloss}.

The proposed \textbf{Lo}w-rank \textbf{R}escoring for \textbf{B}ERT (LoRB) is evaluated on both a public dataset and internal datasets covering a range of domains. We show that \textbf{LoRB} can achieve comparable performance on the target domain and even better performance on non-target domains, as compared to full fine-tuning and other parameter-efficient methods, using only \textbf{0.08\%} of the trainable parameters updated in fine-tuning.  Additionally, LoRB can save up to \textbf{32\%} training memory utilization and achieve up to \textbf{6-fold} reduction in training times, by allowing training with a larger learning rate.

\begin{figure*}[htb]
\centering
    \includegraphics[width = 0.8\textwidth]{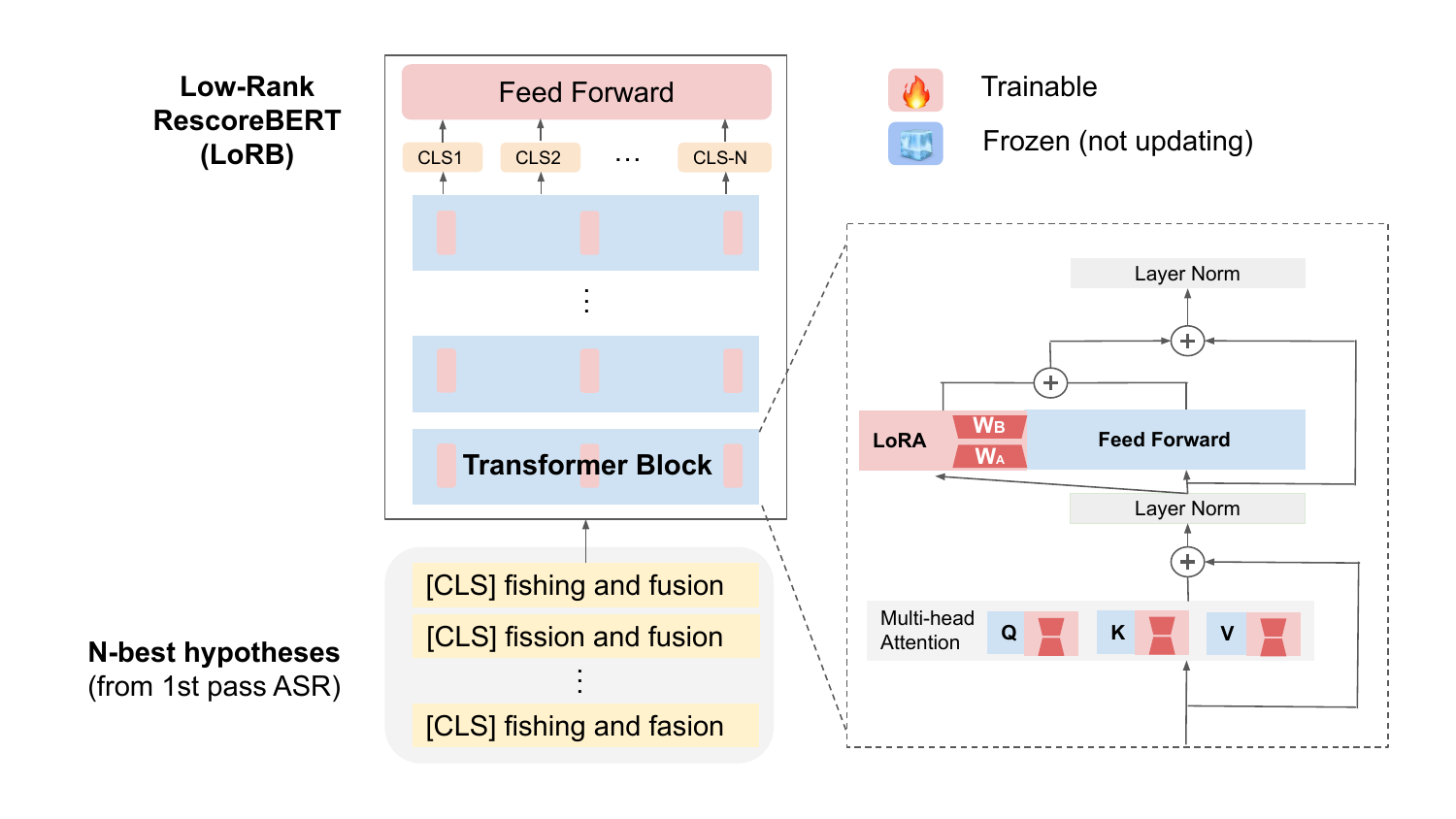}
	\caption{Illustration of the Low-Rank adaptation based Rescoring BERT (LoRB).}
	\label{fig:LoRB}
\end{figure*}
\label{sec:intro}
\section{Related work}
\subsection{Low-rank adaptation}

LoRA has been widely investigated in the natural language processing (NLP) domain. For example, ~\cite{valipour2022dylora} explores an automatic way to select the optimal rank value of LoRA matrices. ~\cite{pu2023empirical, lawton2023neural} discuss the most effective transformer modules in which to insert LoRA matrices, while~\cite{zhang2023adaptive} examines the parameter allocation among weight matrices. Some studies have investigated the underlying reasons for the effectiveness of LoRA. \cite{fu2022effectiveness, chen2023chapter} discovered that the sparsity of learned weights imposes a regularization effect on the original model, resulting in improved generalization. \cite{he2022parameter} demonstrated that constraining the dimensionality of the optimization problem can effectively mitigate catastrophic forgetting. Beyond NLP, low-rank adaptation has also been applied in vision tasks by fine-tuning of vision transformers~\cite{he2022parameter,chavan2023one,kothari2023motion}. However, it remains to be seen whether the findings for NLP and vision tasks can be transferred to second-pass rescoring in automatic speech recognition.

\subsection{Domain adaptation for ASR}

In the domain adaptation research for ASR, the focus has been largely on first-pass acoustic models. Strategies such as contextual biasing have been widely used for RNN-T models~\cite {choudhury2022likelihood, pandey2023procter}. Additionally, for low-resource target domains, self-supervised training and semi-supervised training strategies have been explored~\cite{hwang2022large, chen2023estimate, yang2021voice2series} using speech model reprogramming or adapters. 

For second-pass models, \cite{liu2021domain} explored fine-tuning a general rescoring model for new domains and incorporating a domain classifier to switch between domain-specific models.  \cite{dingliwal2021domain} proposed training of prompt embeddings for target domains and attaching them to the N-best list before scoring with the rescoring GPT2 model. However, this method introduces additional inference latency due to the prepended prompts. Our work, by contrast, aims to explore the generalization effects of low-rank parameter-efficient fine-tuning methods, while reducing the computational cost of domain adaptation without introducing additional inference latency.

\label{sec:relevant_work}
\section{Approach}

\subsection{Discriminative training for second-pass rescoring}\label{sec:mwer}
\subsubsection{Second-pass rescoring}
In this section, we formulate the second-pass rescoring task. Given an \textit{N}-best hypothesis list $E = \{E_1, E_2, \ldots, E_n\}$ obtained from the beam search in the decoder based on the first-pass acoustic model, the rescoring model will generate scores for each hypothesis. For any hypothesis $E_i \in E$, denote by $s^a_i$ the score given by the first pass, and by $s^l_i$ the score produced by the second pass. For both passes, the score of a hypothesis represents the negative log likelihood, thus a lower score represents a more likely hypothesis. 

The language model, such as BERT, takes a hypothesis and outputs a hidden representation $g_i$, then the feed-forward network takes the representation of the task-specific [CLS] token as input and derives the second-pass score $s^l_i$, as shown by Equation~(\ref{eq:rescoring}):
\begin{equation}
    g_i = \text{BERT}(E_i)
\end{equation}
\begin{equation}
    s^l_i = \text{FFNN}(g_i^\mathrm{CLS})
    \label{eq:rescoring}
\end{equation}

The final score of a hypothesis is the linear combination of the first- and second-pass scores:
\begin{equation}
    s_i = s^a_i + \beta \cdot s^l_i
\end{equation}

\subsubsection{Discriminative training objective}
Discriminative training has been widely explored for second-pass rescoring. Specifically, BERT as a masked language model has been applied to second-pass rescoring \cite{xu2022rescorebert} by training with a discriminative objective of minimum word error rate (MWER) \cite{prabhavalkar2018minimum}. Given a hypothesis $E_i \in E$, denote by $\epsilon_i$ the number of word errors (edit distance) from the ground truth transcription. The MWER loss function is defined as the expected number of word errors for the N-best hypothesis, as shown by Equation~(\ref{eq:mwer}):
\begin{equation}
    P_i = \frac{e^{-s_i}}{\sum_{j=1}^n e^{-s_j}}
\end{equation}
\begin{equation}
    \bar{\epsilon}_H = \frac{1}{n} \sum_{i=1}^n \epsilon_i
\end{equation}
\begin{equation}
   \mathcal{L}_\mathrm{MWER} = \sum_{i=1}^n P_i \cdot (\epsilon_i - \bar{\epsilon}_H)
   \label{eq:mwer}
\end{equation}

\subsection{Low-rank adaptation to ASR rescoring} \label{sec:lora}

In the previous modification of BERT for the rescoring task, the pretrained weights $\Phi_0$ of BERT are updated to $\Phi_0 + \Delta \Phi$ by following the gradient for minimizing the MWER loss. The process of learning task-relevant parameters $\Delta \Phi$ is known as the full fine-tuning process. In the full fine-tuning process, the dimension of the learned parameters $|\Delta \Phi|$ equals that of the pretrained weights $|\Phi_0|$.

As shown by \cite{aghajanyan2020intrinsic}, pretrained language models have a low intrinsic dimension and can learn efficiently through a low-dimensional reparameterization. Inspired by this finding and the success of low-rank adaptation of large language models in NLP tasks \cite{hu2021lora}, we propose adapting BERT for the rescoring task by learning a low-rank representation $\Theta$ that has a much smaller dimension than $\Phi_0$, or $|\Theta|  \ll |\Phi_0|$.

Formally, for any dense layer in the transformer blocks with input $x$ and output $h$, denote the pretrained weight as $W_0 \in \mathbb{R}^{d \times k}$, and the updates to the weight as $\Delta W$.
We perform a low-rank decomposition to the updates $\Delta W = W_BW_A$, where $W_B \in \mathbb{R}^{d \times r}$, $W_A \in \mathbb{R}^{r \times k}$ and $r \ll \min(d,k)$. The forward pass is modified to be
\begin{equation}
    h = W_0x + \Delta Wx = W_0x + W_B W_Ax
\end{equation}
During training, $W_0$ is frozen and only $W_A$ and $W_B$ are updated. In BERT, LoRA can be applied to any subset of weight matrices, for example, $W_0$ could be $W_q$, $W_k$, $W_v$ or $W_o$ inside a self-attention module, or be the weight matrices in the two-layer feed-forward network, i.e., $W_{f_1}$ and $W_{f_2}$.

\subsection{Multi-loss training with regularization}\label{sec:multiloss}
Fine-tuning large pretrained models often leads to overfitting on the training data for downstream tasks \cite{jiang2019smart,aghajanyan2020better}. Even though some parameter-efficient fine-tuning methods are shown to be helpful in alleviating the overfitting issues by constraining the number of trainable parameters \cite{xu2022evaluating, yang2023parameter, chen2023exploring}, in some of our experiments a marginal degradation of performance on unseen test sets is observed when evaluating the LoRA fine-tuned rescoring model.

In order to obtain a hidden representation from the pretrained BERT with better generalization performance, we add a correlation-based regularization loss $\mathcal{L}_{cor}$ besides the MWER loss:
\begin{equation}
    \mathcal{L} = \mathcal{L}_\mathrm{MWER} + \lambda \mathcal{L}_{cor}
\end{equation}

The correlation-based regularization~\cite{zhang2022fine} has been proposed to alleviate the representation degeneration~\cite{gao2019representation} problem caused by fine-tuning on pretrained language models. By forcing the feature space of representations to be more isotropic (uniformly variable in all directions), the expressiveness of the learned representation can be preserved better. Formally, the correlation-based regularization loss is defined so as to penalize the correlation matrix for sentence representations for deviating from the identity:
\begin{equation}
   \mathcal{L}_{cor} = \lVert\mathrm{\Sigma} - \mathrm{I}\lVert
\end{equation}
where $\lVert \cdot \lVert$ denotes the Frobenius norm, $\mathrm{I} \in \mathbb{R}^{d_h \times d_h}$ is the
identity matrix, $\mathrm{\Sigma} \in \mathbb{R}^{d_h\times d_h}$ is the correlation matrix
with $\Sigma_{ij}$ being the Pearson correlation coefficient
between the $i$th dimension and the $j$th dimension of the hidden representation of the [CLS] token $g^\mathrm{CLS} \in \mathbb{R}^{d_h}$. In the case of LoRB, only the LoRA matrices that contribute to the hidden representation of the [CLS] token in each BERT layer are regularized by the correlation-matrix loss.

\begin{table*}[!h]
\centering
\caption{Relative WER improvement of LoRB, full fine-tuning (FT), Adapter and BitFit when fine-tuning on messaging data.} 
\label{tab:msg_wer}
\begin{tabular}{l|r|r|r|r|r}
\toprule
  & & \multicolumn{1}{c|}{Target Domain}  &\multicolumn{2}{c}{Non-Target Domain }      &           \\\hline
    \multirow{2}{*}{Method} &\% Trainable &\multirow{2}{*}{Messaging$_{\text{Test}}$} & \multirow{2}{*}{General} &\multirow{2}{*}{Shopping}  & \multirow{2}{*}{Knowledge} \\
    &Parameters & & &\\
    \midrule
    RescoreBERT$_\text{pretrained 170M}$& non-adapted &baseline &baseline &baseline  &baseline \\\hline
    w/ Fine-Tuning (FT)& 100\% & 3.30\% &-2.33\%  &-1.17\% &-0.34\%\\
    w/ Residual Adapter &1.27\% &3.72\% &-16.60\% &-17.33\% &-17.07\%\\
    w/ BitFit & 0.01\% &3.30\%  &-18.83\% &-17.57\% &-20.90\%\\
    w/ Prefix & 0.05\% &3.30\%  &-1.98\% &-1.53\% & -1.39\%\\ \hline
    LoRB & 0.08\% & \cellcolor[HTML]{9AFF99}\textbf{6.06\%} &\cellcolor[HTML]{9AFF99}\textbf{0.27\%}&\cellcolor[HTML]{9AFF99}0.23\% &\cellcolor[HTML]{9AFF99}\textbf{0.34\%}\\
    LoRB + $\mathcal{L}_{cor}$ & 0.08\% &\cellcolor[HTML]{9AFF99}5.65\% & \cellcolor[HTML]{9AFF99}-0.51\%&\cellcolor[HTML]{9AFF99}\textbf{0.82\%} &\cellcolor[HTML]{9AFF99}0.01\%\\
    \bottomrule
\end{tabular}
\end{table*}

\section{Experiments}
\begin{figure}[t]
    \includegraphics[scale=0.6]{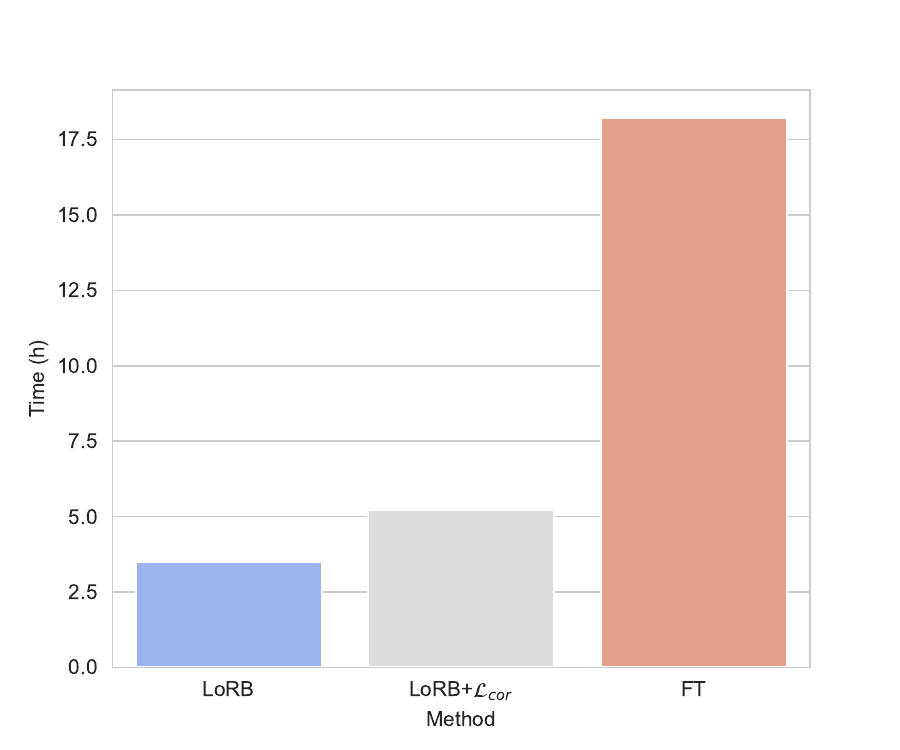}
	\caption{Wall-clock training time of LoRB, LoRB+$\mathcal{L}_{cor}$ and Fine-Tuning (FT) when training on \textit{messaging} data.
	}
	\label{fig:lr_and_time}
\end{figure}

\subsection{Datasets}
The training datasets for domain adaptation include one public dataset, LibriSpeech \cite{panayotov2015librispeech}, and two internal datasets: \textit{Messaging} (350 hours) and \textit{Music} (150 hours). Furthermore, we explore the scaling behavior with regard to the sizes of the pretrained model and the training data, using an internal \textit{conversational domain} dataset.

We evaluate the low-rank adaptation of the language model on three internal datasets drawn from from de-identified, far-field English-language conversations with a voice assistant. The internal \textit{General} domain set contains 194 hours, the \textit{Shopping} domain set contains 20 hours, and the \textit{Knowledge} domain set contains 5 hours of training data, respectively.

\subsection{Implementation}

In the adaptation experiments, we vary the LoRA rank over the values \{4,8,16,32\} and apply LoRA to two sets of target modules: [$W_q$, $W_v$] and [$W_q$, $W_k$, $W_v$, $W_{f_1}$, $W_{f_2}$]. In the LoRA layer, we set the dropout rate to $0.01$ and $\alpha=32$. When fine-tuning RescoreBERT, we initialize the feed-forward network in RescoreBERT from the pretrained model checkpoints and continuously update the parameters in the feed-forward network, as shown in Figure~\ref{fig:LoRB}. For all parameter-efficient training methods and full fine-tuning, we use early stopping to evaluate the checkpoint with best performance on an in-domain validation set.

For LibriSpeech, we fine-tune the cased BERT$_\text{base}$ model for fair comparison with previous work. For other internal training datasets, we fine-tune an in-house 170M  RescoreBERT model with 16 layers and 1024-dimensional hidden
layers, which was trained on internal data with the discriminative training objective for 435K steps.

\subsection{Baselines}
The word error rate (WER) of the first-pass RNN-Transducer speech recognition baseline system used is below 10\%. We compare the fine-tuning results of low-rank adaptation with full fine-tuning and three other parameter-efficient fine-tuning methods. Here the ``Adapter'' method refers to the standard residual adapter proposed in \cite{houlsby2019parameter}, which has a latent dimension that is half of its encoder dimension, $768$. Adapter layers are inserted into the self-attention module and the subsequent residual connection, as well as into the MLP module and its subsequent residual connection. Each adapter layer includes two fully connected layers, bias vectors, and a nonlinearity placed between them. The ``BitFit'' method, proposed in \cite{zaken2021bitfit}, involves training the bias vectors in each module while freezing all other parameters. The ``Prefix'' method refers to prefix-tuning~\cite{li2021prefix}, which inserts trainable tokens into input sequence.

\section{Results and analysis}
\subsection{Low-rank domain adaptation}

\subsubsection{Messaging data as continuous domain adaptation}

Table \ref{tab:msg_wer} shows the evaluation results on four internal datasets. We fine-tune a 170M RescoreBERT model with the MWER training objective on an internal \textit{messaging} (MSG) dataset. The fine-tuned models are evaluated on both in-domain \textit{messaging} test set and out-of-distribution data from the \textit{General}, \textit{Shopping} and \textit{Knowledge} domains. The first row shows the test evaluation results of the 170M RescoreBERT model without any fine-tuning. All parameter-efficient fine-tuning methods achieves performance comparable to or better than full fine-tuning (FT) on the target domain \textit{Messaging}. However, FT, Adapter and BitFit suffer from performance degradation on out-of-distribution data, while LoRB performs robustly in both target domain and nontarget domains. %

\subsubsection{Case Study 1: Effect of regularization}
Table \ref{tab:hotfix_wer} presents the performance comparison of LoRB and LoRB with correlation-based regularization against baseline methods on three internal test sets from nontarget domains. Our experiments reveal that the Music domain data is prone to overfitting when fine-tuning is applied, resulting in degradation on other domain data. This can be attributed to the limited dataset size and the presence of challenging rare words like artist names. While both Adapter and LoRB techniques exhibit some level of improvement in mitigating the degradation across most domains, the combination of LoRB with correlation-based regularization results in the most substantial improvement in performance.%

\begin{table}[!h]
\caption{Relative WER improvement of LoRB$_{170M}$, full fine-tuning (FT) and Adapter when fine-tuning on Music data.} 
\label{tab:hotfix_wer}
\adjustbox{scale=0.85}{
\centering
\begin{tabular}{l|c|c|c|c}
\toprule
    & \multicolumn{4}{l}{Non-Target}            \\\hline
    Method & General& Shopping & Knowledge & Average \\\midrule
    
    Fine-Tuning (FT)  &baseline &baseline & baseline & baseline  \\
    Residual Adapter  &-0.14\% &0.49\% &0.3\% & 0.22\%  \\ \hline
    LoRB$_{170M}$&-0.5\% &0.21\% &0.90\% & 0.20\%  \\
    LoRB$_{170M}$ + $\mathcal{L}_{cor}$ &\textbf{0.22\%} &\textbf{0.71\%} & \textbf{1.21\%} & \textbf{0.71\%}\\
    \bottomrule
\end{tabular}}

\end{table}

\subsubsection{Case Study 2: Public dataset}

Table~\ref{tab:librispeech_wer}
shows the WER on test-Clean and test-Other portions of the LibriSpeech dataset. We follow a Whisper setup~\cite{shivakumar2023distillation} for first-pass decoding. On both test sets, LoRB achieves the largest reduction in WER compared to other parameter-efficient training methods.  Specifically, in test-Other, LoRB can achieve results comparable to FT with only 0.27\% of the parameters, and the correlation-based loss brings further improvements, which aligns with our findings in Case Study 1.

\begin{table}[!h]
\caption{Absolute WER on the two standard test sets of public LibriSpeech~\cite{panayotov2015librispeech} baseline decoded by Whisper-tiny. The 170M BERT base model is retrieved from official public release~\cite{devlin2019bert} for reproducible evaluation under Apache License.} 
\label{tab:librispeech_wer}

\adjustbox{scale=0.9}{
\centering
\begin{tabular}{l|c|c|c}
\toprule
    Model \& Method  &\% Params & test-Clean & test-Other \\\midrule
    BERT$_\text{base-cased}$& non-adapted &6.17 & 13.81 \\\hline
    w/ FT& 100\% &\textbf{4.37} & 10.80 \\ \hline
    w/ Residual Adapter& 2.15\%&5.29  &12.01  \\
    w/ BitFit& \textbf{0.01}\%& 5.60 &12.43  \\
    w/ Prefix& 0.34\%&5.30  &12.05  \\ \hline
    LoRB$_{170M}$&0.27\% &\cellcolor[HTML]{9AFF99}4.50 &\cellcolor[HTML]{9AFF99}10.81  \\
    LoRB$_{170M}$ + $\mathcal{L}_{cor}$&0.27\% &\cellcolor[HTML]{9AFF99}\textbf{4.47} &\cellcolor[HTML]{9AFF99}\textbf{10.78}  \\
    \bottomrule
\end{tabular}}
\end{table}

\subsubsection{Analysis: Training stability}

Table~\ref{tab:msg_stability} shows the word error rate after full fine-tuning and LoRB under different training hyper-parameter settings. We observed that FT is brittle for various combinations of warm-up steps and learning rate schedules, while LoRB is more robust to changes in hyperparameters.

\subsubsection{Analysis: Training time and GPU memory utilization}

A training time comparison is shown in Figure~\ref{fig:lr_and_time}. We find that, while LoRB takes longer to converge compared to FT at the same learning rate, the performance of FT degrades greatly when the learning rate is increased. As a result, we can utilize LoRB to achieve a similar WER as FT with shorter training time by benefiting from the larger learning rate, as shown in Figure~\ref{fig:lr_and_time}. Furthermore, we find that LoRB can reduce the GPU memory percentage used during training substantially, from 87\% to 52\%.

\begin{table}[!th]
\caption{Relative WER improvement on nontarget Shopping domain compared to  170M RescoreBERT without fine-tuning, under different warm-up steps and learning rate combinations.} 
\label{tab:msg_stability}

\adjustbox{scale=0.95}{
\centering
\begin{tabular}{l|c|c|c|c}
\toprule
    &\multicolumn{4}{c}{WER} \\\cline{1-5} 
    & \multicolumn{2}{c|}{warmup=5k} &
    \multicolumn{2}{c}{warmup=10k} \\\midrule
    & lr=$1\text{e-}5$& lr=$1\text{e-}7$& lr=$1\text{e-}5$&
     lr=$1\text{e-}7$ \\\midrule
     RescoreBERT&baseline &baseline &baseline & baseline \\
    FT&\cellcolor[HTML]{FFCCC9}-72.2\% &\cellcolor[HTML]{FFCCC9}-2.0\% &\cellcolor[HTML]{FFCCC9}-6.48\% & \cellcolor[HTML]{FFCCC9}-1.17\% \\
    LoRB$_{170M}$&0 &0 &\cellcolor[HTML]{9AFF99}+0.23\% &\cellcolor[HTML]{9AFF99}+0.11\%   \\
    \bottomrule
\end{tabular}}
\end{table}

\begin{figure}[h]
\centering
    \includegraphics[scale=.5]{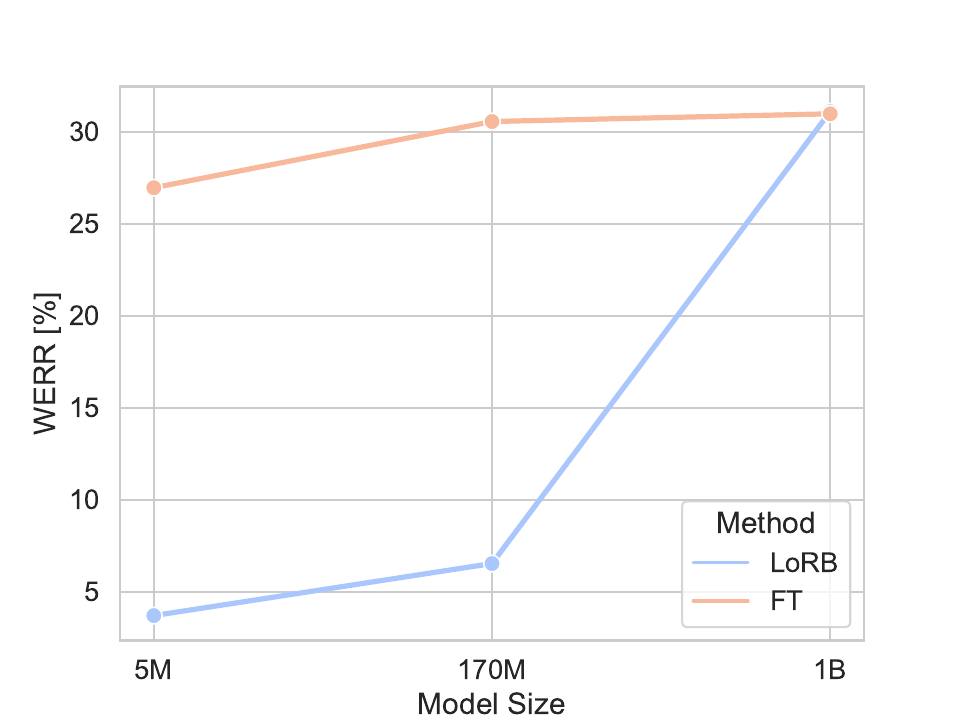}
	\caption{WER on a conversational test set evaluated by RescoreBERT of size 5M, 170M and 1B, fine-tuned with ``conversational domain'' data using FT and LoRA.}
	\label{fig:scale_model}
\end{figure}

\subsubsection{LLM scaling results}

In this section, we show how the scale of the underlying pretrained language model and the scale of the training dataset can affect the performance of LoRB. We use an internal conversational dataset (roughly 60M utterances) as the training source. To evaluate the scaling behavior for varying pretrained model sizes, we fine-tune in-house RescoreBERT models with 5M, 170M and 1B parameters, respectively, on a set of 150K conversational training utterances. To investigate the scaling behavior for data sizes, we split the conversational training data into five log scales with roughly 20M/5M/1500K/500K/150K utterances, respectively.

Figure~\ref{fig:scale_model} shows the scaling with regard to model size. With the size of the pretrained language model increasing, the performance gap between FT and LoRB shrinks. With the increase in total pretrained parameters of the backbone model, the performance gap between FT and LoRB is reduced from -22.3\% (at the scale of 170M) to +2.4\% (at the 1B scale) in terms of WER relative (WERR) difference. In our ASR rescoring model experiments, we found
that a larger BERT model size improves the convergence speed of LoRB by a factor of 2.74, which has benefits for production-size deployments.

 \begin{figure}[h]
 \centering
     \includegraphics[scale=.48]{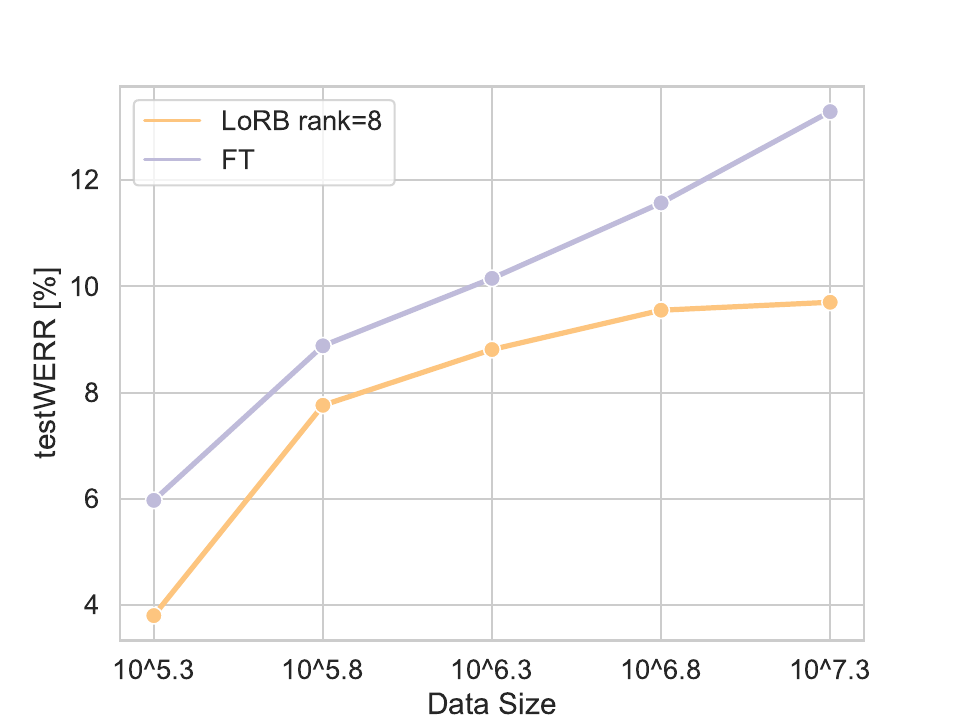}
 	\caption{WER evaluated by 1B RescoreBERT, fine-tuned with various sizes of ``conversational domain'' data using FT and LoRA.}
 	\label{fig:scale_data}
 \end{figure}

  \begin{figure}[h!]
 \centering
     \includegraphics[scale=.48]{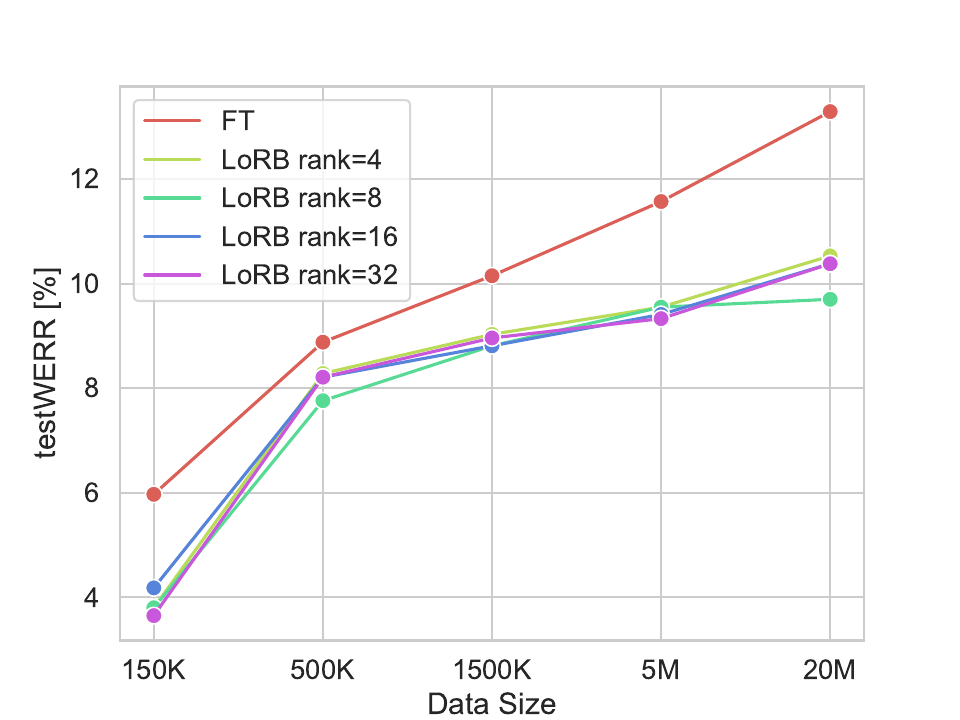}
 	\caption{WER as a function of data size, evaluated by 1B RescoreBERT, fine-tuned with FT and various ranks of LoRA.}
 	\label{fig:scale_data_allrank}
 \end{figure}
 Figure~\ref{fig:scale_data} shows the WER on the same conversational test set for models trained on different amount of data. In general, we observe that a larger data size correlates with greater improvement in performance. Notably, the improvement resulting from a change in data scale from $150K$ to $500K$ is nearly four times that observed when transitioning from $500K$ to $20M$ for LoRB. Unlike the linear scaling law observed in full fine-tuning ~\cite{gu2023scaling}, LoRB follows a logarithmic scaling curve, approaching a fixed value as the data size reaches a certain threshold. Figure~\ref{fig:scale_data_allrank} shows the scaling of LoRB across various rank sizes. While there is no obvious correlation between rank value and word error rate across different data scale settings, the general trend remains consistent: larger dataset sizes lead to a more substantial performance gap compared to full fine-tuning (FT).

\section{Conclusion}
We have introduced LoRB, an efficient and scalable low-rank decomposition for domain-adaptation of BERT-based rescoring models with low computation cost and no performance degradation when trained on limited-size in-domain data. By inserting weight matrices amounting to only $0.08$\% of the parameters of the pretrained models and freezing all other parameters, we achieve speech recognition performance comparable to full fine-tuning with a 6-fold speedup in training.  Experimental rescoring results on public and internal datasets demonstrate the effectiveness and generalization of the LoRB framework and a correlation-based multi-loss training. The scaling results highlight the importance of large pretrained models for best speech recognition rescoring results.

\clearpage
\bibliographystyle{IEEEbib}
\small
\bibliography{strings,refs}

\end{document}